\documentclass{article}
\usepackage{spconf,amsmath,graphicx,booktabs,array,multirow,siunitx,amssymb,subcaption,float,afterpage}

\title{From Movements to Metrics: Evaluating Explainable AI Methods in Skeleton-Based Human Activity Recognition}
\name{Kimji N. Pellano$^{1}$, Inga Strümke$^{2}$, and Espen Alexander F. Ihlen$^{1}$} 
\address{
\begin{tabular}{@{}l@{}}
\small $^{1}$ Department of Neuromedicine and Movement Science, Faculty of Medicine and Health Sciences, \\
\small Norwegian University of Science and Technology, 7034 Trondheim, Norway\\
\small$^{2}$ Department of Computer Science, Faculty of Information Technology and Electrical Engineering, \\
\small Norwegian University of Science and Technology, 7034 Trondheim, Norway
\end{tabular}
}

\begin{document}
%
\maketitle
\begin{abstract}
The advancement of deep learning in human activity recognition (HAR) using 3D skeleton data is critical for applications in healthcare, security, sports, and human-computer interaction. This paper tackles a well-known gap in the field, which is the lack of testing in the applicability and reliability of XAI evaluation metrics in the skeleton-based HAR domain. We have tested established XAI metrics namely \textit{faithfulness} and \textit{stability} on Class Activation Mapping (CAM) and Gradient-weighted Class Activation Mapping (Grad-CAM) to address this problem. The study also introduces a perturbation method that respects human biomechanical constraints to ensure realistic variations in human movement. Our findings indicate that \textit{faithfulness} may not be a reliable metric in certain contexts, such as with the EfficientGCN model. Conversely, \textit{stability} emerges as a more dependable metric when there is slight input data perturbations. CAM and Grad-CAM are also found to produce almost identical explanations, leading to very similar XAI metric performance. This calls for the need for more diversified metrics and new XAI methods applied in skeleton-based HAR.
\end{abstract}
\section{Introduction}

\label{sec:introduction}
Analyzing human movement through 3D skeleton data has promising non-trivial applications in high-stake sectors such as in healthcare and rehabilitation \cite{nguyen2016skeleton}, security and surveillance \cite{liu2021self}, sports and athletics \cite{guo2021attention}, and human-computer interaction \cite{usman2022skeleton}. Because of this, integrating deep learning in skeleton data analysis requires an understanding of the model's decision-making processes. One particular application is compliance with the EU's proposed AI Act which emphasizes that transparency and human oversight should be embedded in high-risk applications such as AI-assisted medical diagnostics \cite{eu2021aiact}. A basic form of deep learning technique applied to human movement analysis using skeleton data is human activity/action recognition (HAR). State-of-the-art HAR models are continually being developed and improved. It started with the introduction of Spatial Temporal Graph Convolutional Networks (ST-GCN) architecture in 2018 \cite{yan2018spatial} as it was the first to use graph convolution for HAR. ST-GCN then became the baseline for dozens of emerging skeleton-based HAR models that seek to improve this original implementation. 

Recent advancements in HAR model architectures have been significant, but strides in their explainability remain limited. Class Activation Mapping (CAM) was used in EfficientGCN \cite{song2020stronger} and ST-GCN \cite{ghaleb2021skeleton} to highlight the body points significant for specific actions. In \cite{das2022gradient}, Gradient-weighted Class Activation Mapping (Grad-CAM) was implemented in ST-GCN. There is a growing trend towards using explainable AI (XAI) methods, extending from CNNs to ST-GCNs, yet XAI metrics to assess their reliability in this domain have yet to be tested. There is also a lack of comparative analysis between these methods, which leaves a gap in understanding their relative performance in HAR applications. Additionally, research is limited on metrics that assess XAI methods in the context of data perturbation.

While the paper in \cite{das2022gradient} evaluated the \textit{faithfulness} and \textit{contrastivity} of Grad-CAM, it did not offer insights into its performance relative to other XAI methods. Moreover, their choice of using masking/occlusion to check for changes in prediction output raises concerns. Masking can potentially distort the standard human skeletal structure that GCN-based models are trained on. Movements of the human body are governed by biomechanical principles, and perturbations that do not respect these principles can potentially result to misleading understanding of the model's faithfulness. Recognizing the growing relevance of skeleton-based HAR models in critical areas, our paper aims to test established metrics that assess their corresponding feature attribution techniques. Additionally, no other study has assessed explainability metrics using biomechanically accurate perturbations of the skeletal graph. In \cite{wang2021understanding} and  \cite{liu2020adversarial}, perturbation of skeleton data was employed, which they claim to have maintained normal human kinematics during perturbation. However, the objective was for adversarial attack thus the perturbed skeleton joints were neither controlled nor deliberately targeted. In essence, we also tackle the absence of evaluation metrics grounded in biomechanically correct perturbations of the skeletal graph.

Alongside the pursuit of improved explainability, assessing the stability of model decisions and their explanations is important. As human skeletal data can exhibit subtle variances due to minor changes in posture, movement, or data capturing techniques, the decisions from the model and the explanations from the XAI methods should remain consistent and trustworthy. That is, dramatic shifts in explanations due to minor input changes cast doubt on model reliability. Moreover, imprecise estimation of joint center positions in 3D skeletal data analysis underscores the need to evaluate decision and explanation robustness using biomechanically and kinematically correct perturbations. To address this, we draw from metrics established for other data types. In this work, we focus on the two primary metrics highlighted in \cite{agarwal2022openxai}: \textit{faithfulness} \cite{zhou2021evaluating}, which gauges how closely an explanation mirrors the model's internal logic and \textit{stability} \cite{alvarez2018robustness}, which pertains to the consistency of a model's explanations across similar inputs.

This paper's key contributions are:
\begin{itemize}
    \item Testing established metrics and assessing its applicability for evaluating XAI methods in skeleton-based HAR.
    \item Introducing a controlled approach to perturb 3D skeleton data for XAI evaluation, ensuring biomechanically correct variations for realistic skeletal data representation.
    \item Assessing the impact of perturbation magnitude variations on metrics.
    \item Comparing the performance of CAM, Grad-CAM, and a random feature attribution method for HAR.
\end{itemize}
\section{Materials}
To provide the framework for this research, the dataset used, the neural network architecture trained and tested, and the XAI metrics implemented are briefly summarized below.
\subsection{NTU RGB+D 60 dataset and EfficientGCN}

The NTU RGB+D-60 dataset \cite{shahroudy2016ntu} contains 60 action classes with over 56 thousand 3D skeleton data, each composed of sequential frames captured from 40 different subjects using the Kinect v2 camera with depth sensor. For evaluation purposes, the dataset is further divided into cross-subject and cross-view subgroups – the former is composed of different human subjects performing the same actions, while the latter uses different camera angle views.
\begin{figure}[htb]
  \centering
  \includegraphics[width=7cm]{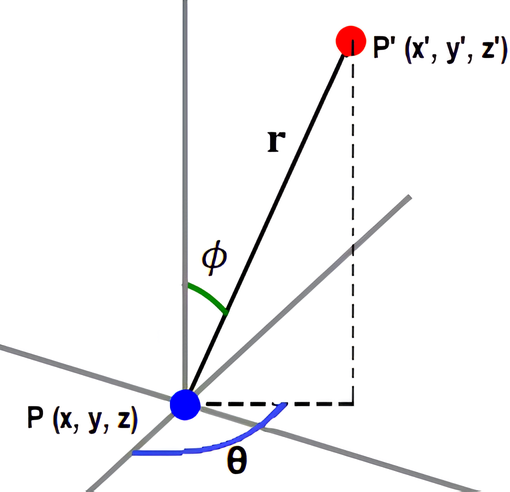}
  \caption{Illustration of perturbing a point P(x, y, z) in 3D space to a new position P\textquotesingle(x\textquotesingle, y\textquotesingle, z\textquotesingle) using spherical coordinates. The perturbation magnitude is represented by $r$, with azimuthal angle $\theta$ and polar angle $\phi$.}
  \label{fig:spherical}
\end{figure}

The EfficientGCN architecture \cite{song2022constructing} is a result of extending the concept of EfficientNet \cite{tan2019efficientnet} for CNNs into ST-GCNs to reduce the computing resource demand for HAR. There are a total of 24 different EfficientGCN network configurations with different scaling coefficients that the user can choose and test. In this paper, we use the B4 configuration which has achieved the highest accuracy at 92.1\% on cross-subject dataset and 96.1\% on cross-view, compared to 81.5\% and 88.3\% of the baseline ST-GCN, respectively. 
\begin{figure*}[htb]
  \centering
  \includegraphics[width=13cm]{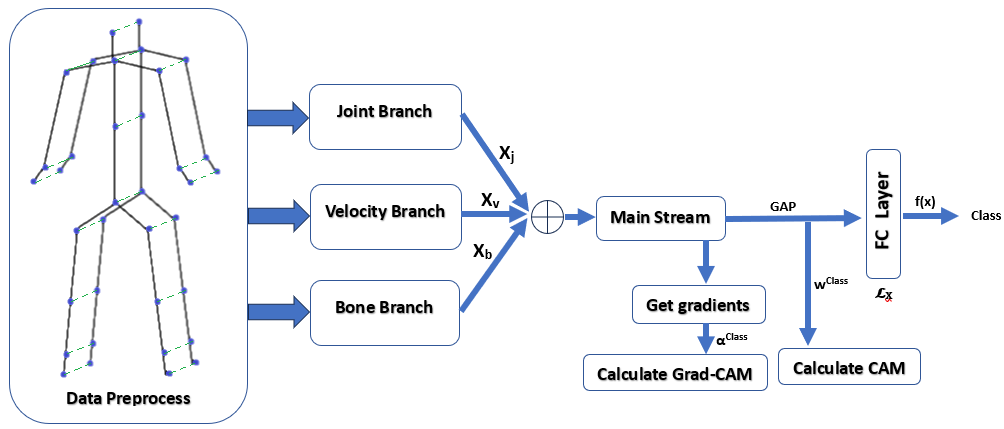}
  \caption{The EfficientGCN pipeline \cite{song2022constructing} showing the variables for calculating \textit{faithfulness} and \textit{stability}. Perturbation is performed in \textit{Data Preprocess} stage.}
  \label{fig:flow}
\end{figure*}
\subsection{Evaluation Metrics}

\subsubsection{Faithfulness}

Predictive faithfulness refers to the degree to which the changes in an explanation's features meaningfully alter the model's prediction, indicating the explanation's alignment with the model's actual reasoning process. Prediction Gap on Important feature perturbation (PGI) measures how much prediction changes when top-$k$ features are perturbed. Conversely, Prediction Gap on Unimportant feature perturbation (PGU), measures the change in prediction when unimportant features are perturbed. The two metrics are derived from the formulas below \cite{agarwal2022openxai}. Let $\mathbf{X}$ represent the original input data with its associated explanation $e_{\mathbf{X}}$, and $f(\mathbf{\cdot})$ the output probability. Then, $\mathbf{X}'$ signifies the perturbed variant of $\mathbf{X}$, and $e_{\mathbf{X}'}$ the revised explanation after perturbation.

\begin{equation}
PGI(\mathbf{X}, f, e_{\mathbf{X}}, k) = \mathbb{E}_{\mathbf{X}' \sim \text{perturb}(\mathbf{X}, e_{\mathbf{X}}, \text{top-}k)}[|f(\mathbf{X}) - f(\mathbf{X}')|]
\label{eq:pgi}
\end{equation}

\begin{equation}
PGU(\mathbf{X}, f, e_{\mathbf{X}}, k) = \mathbb{E}_{\mathbf{X}' \sim \text{perturb}(\mathbf{X}, e_{\mathbf{X}}, \text{non top-}k)}[|f(\mathbf{X}) - f(\mathbf{X}')|]
\label{eq:pgu}
\end{equation}

\subsubsection{Stability}
The concept of stability refers to the maximum amount of change in explanation (i.e. attribution scores) when the input data is slightly perturbed. There are three sub-metrics that can be calculated.

Relative Input Stability (RIS) measures the maximum change in attribution scores with respect to a corresponding perturbation in the input. Given that EfficientGCN has multiple input branches, it is essential to compute the RIS for each branch namely joint, velocity, and bone. From hereon, they shall be referred to as RISj, RISv, and RISb, respectively.
\begin{equation}
\begin{split}
\operatorname{RIS}\left(\mathbf{X}, \mathbf{X}^{\prime}, \mathbf{e}_{\mathbf{X}}, \mathbf{e}_{\mathbf{X}^{\prime}}\right) &= \max_{\mathbf{X}^{\prime}} \frac{\left\|\frac{\left(\mathbf{e}_{\mathbf{X}}-\mathbf{e}_{\mathbf{X}^{\prime}}\right)}{\mathbf{e}_{\mathbf{X}}}\right\|_p}{\max \left(\left\|\frac{\left(\mathbf{X}-\mathbf{X}^{\prime}\right)}{\mathbf{X}}\right\|_p, \epsilon_{\min }\right)}, \\
&\quad \forall \mathbf{X}^{\prime} \text{ s.t. } \mathbf{X}^{\prime} \in \mathcal{N}_{\mathbf{X}} ; f(\mathbf{X}) = f(\mathbf{X}^{\prime})
\end{split}
\label{eq:ris}
\end{equation}
Relative Output Stability (ROS) measures the maximum ratio of how much the explanation changes to how much the model's prediction probability changes due to small perturbations in the input data.
\begin{equation}
\begin{split}
\operatorname{ROS}\left(\mathbf{X}, \mathbf{X}^{\prime}, \mathbf{e}_{\mathbf{X}}, \mathbf{e}_{\mathbf{X}^{\prime}}\right) &= \max_{\mathbf{X}^{\prime}} \frac{\left\|\frac{\left(\mathbf{e}_{\mathbf{X}}-\mathbf{e}_{\mathbf{X}^{\prime}}\right)}{\mathbf{e}_{\mathbf{X}}}\right\|_p}{\max \left(\left\|\frac{\left(f(\mathbf{X})-f\left(\mathbf{X}^{\prime}\right)\right)}{f(\mathbf{X})}\right\|_p, \epsilon_{\min }\right)} \\
&\quad \forall \mathbf{X}^{\prime} \text{ s.t. } \mathbf{X}^{\prime} \in \mathcal{N}_{\mathbf{X}} ; f(\mathbf{X}) = f(\mathbf{X}^{\prime})
\end{split}
\label{eq:ros}
\end{equation}
Relative Representation Stability (RRS) measures the maximum change in a model's explanations relative to changes in the model's internal representations brought about by small input perturbations. In this context, the internal representation denoted as $\mathcal{L}_{\mathbf{X}}$ typically refers to an intermediate layer's output in a neural network, capturing the model's understanding of the input data. In our experiment, we extract and use the logits from the layer preceding the softmax function for our computations.
\begin{equation}
\begin{split}
\operatorname{RRS}\left(\mathbf{X}, \mathbf{X}^{\prime}, \mathbf{e}_{\mathbf{X}}, \mathbf{e}_{\mathbf{X}^{\prime}}\right) &= \max_{\mathbf{X}^{\prime}} \frac{\left\|\frac{\left(\mathbf{e}_{\mathbf{X}}-\mathbf{e}_{\mathbf{X}^{\prime}}\right)}{\mathbf{e}_{\mathbf{X}}}\right\|_p}{\max \left(\left\|\frac{\left(\mathcal{L}_{\mathbf{X}}-\mathcal{L}_{\mathbf{X}^{\prime}}\right)}{\mathcal{L}_{\mathbf{X}}}\right\|_p, \epsilon_{\min }\right)} \\
&\quad \forall \mathbf{X}^{\prime} \text{ s.t. } \mathbf{X}^{\prime} \in \mathcal{N}_{\mathbf{X}} ; f(\mathbf{X}) = f(\mathbf{X}^{\prime})
\end{split}
\label{eq:rrs}
\end{equation}
\section{Methods}
\subsection{Skeleton Data Perturbation}
In 3D space, skeleton joints can be perturbed using spherical coordinates by generating random angles $\theta$ and $\phi$ for perturbation direction, sourced from a Gaussian distribution. The magnitude of this perturbation is controlled by radius $r$. In a standard XAI metric test, we recommend to adhere to the principle that $X'$ should be within the neighborhood of $X$ to ensure that inputs remain representative of human kinematics and avoid skewing model predictions. This means constraining the magnitude of $r$, which in our pipeline is initially set to 2.5cm. When it comes to body point tracking, the Kinect v2's tracking error ranges from 1 to 7 cm compared to the gold-standard Vicon system \cite{otte2016accuracy} so a 2.5cm perturbation ensures the perturbed data stays within Kinect's typical accuracy tolerance. However, we also tested the metrics with increasing $r$ (in cm: 2.5, 5, 10, 20, 40, and 80). While this contradicts our initial recommendation, varying the perturbation magnitude would allow us to (a) test the hypothesis that a small perturbation should result in meaningful changes in the prediction, which should be reflected in \textit{faithfulness} results, and (b) observe its effects on the explanations, which should be reflected in \textit{stability} results.

The point P\textquotesingle(x\textquotesingle, y\textquotesingle, z\textquotesingle) in Figure~\ref{fig:spherical} can be calculated using the equations below, which are used to convert a point from spherical to Cartesian (rectangular) coordinates. In these equations, $r$ represents the distance from the two points, $\theta$ denotes the azimuthal angle, and $\phi$ is the polar angle. A fixed random seed was used to generate reproducible angles $\theta$ and $\phi$. The variables $dx$, $dy$, and $dz$ are computed once, and each joint is given its own unique set of these values. When added to the original coordinates across all video frames, a mildly perturbed 3D point is produced. This method ensures that a particular joint undergoes the same random adjustment across all frames, rather than different perturbations in each frame.
\begin{align*}
x' &= x + dx, \quad & dx &= r \sin(\phi) \cos(\theta) \\
y' &= y + dy, \quad & dy &= r \sin(\phi) \sin(\theta) \\
z' &= z + dz, \quad & dz &= r \cos(\phi)
\end{align*}

\subsection{Calculation and Evaluation of XAI Metrics}
\begin{figure}[htb]
  \centering
  \includegraphics[width=8cm]{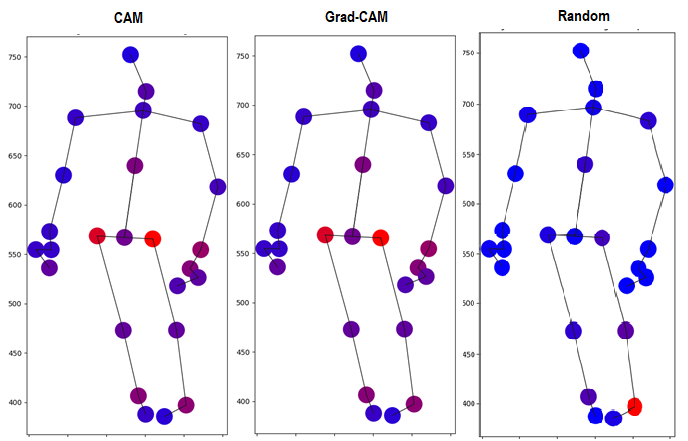}
  \caption{Left to right: CAM, Grad-CAM, and baseline random attributions for a data instance in 'writing' (class 11), averaged for all frames and normalized. The color gradient denotes the score intensity: blue indicates 0, progressing to red which indicates a score of 1.}
  \label{fig:camvsgradcam}
\end{figure}

\begin{equation}
\text{e}_{\text{X, CAM}} = \sum_n w_n^{class} F^n
\label{eq:cam}
\end{equation}
\begin{equation}
\text{e}_{\text{X, Grad-CAM}} = \sum_n \alpha_n^{class} F^n
\label{eq:gradcam}
\end{equation}
Figure~\ref{fig:flow} and Equations~\ref{eq:cam} and~\ref{eq:gradcam} show how to get the variables to calculate the metrics. From the definition of CAM \cite{zhou2016learning}, $w$ in Equation~\ref{eq:cam} are the weights after Global Average Pooling (GAP) for the specific output $class$, and $F^n$ denotes the $nth$ feature map. Similarly, $\alpha$ in Equation~\ref{eq:gradcam} for Grad-CAM \cite{selvaraju2016grad} is calculated by averaging the gradients. Figure~\ref{fig:camvsgradcam} shows a sample CAM and Grad-CAM explanations in comparison with the random baseline. For PGI and PGU, we begin by determining the average attribution scores for each skeletal joint across all frames. Next, the joints are ranked based on their order of significance. The top-$k$ joints, along with their respective non-top-$k$ counterparts, are then perturbed by selecting $k$ nodes according to their attribution score ranking. For each data instance and for values of $k$, PGI and PGU are computed. The results are then aggregated by calculating the AUC, which allows for reporting a single value for each metric. 

Since the metrics are unitless, a random method serves as the benchmark for the least desirable outcome by randomly assigning feature attribution scores. Higher PGI values are optimal, indicating that altering important skeletal nodes has a marked impact on the prediction. Lower PGU values are better, suggesting that perturbing the identified unimportant skeletal nodes do not cause significant change in the model's output prediction. Lastly, a \textit{stability} (RIS, RRS, and ROS) closer to zero is indicative of a model's robustness, signifying that minor perturbations to the input data do not lead to significant changes in the explanation. In order to thoroughly assess the applicability and consistency of the XAI evaluation metrics, we test them on both the most accurately classified class (class 26 - `jump up') and the class with the highest misclassification rate (class 11 - `writing') in the NTU dataset. Of the 276 samples in the class 26 test set, only 1 sample was misclassified by the EfficientGCN-B4 model, while only 174 were correctly classified out of 272 samples in class 11.

\section{Results}
\begin{figure*}
    \begin{subfigure}{0.3\textwidth}
        \includegraphics[width=\textwidth]{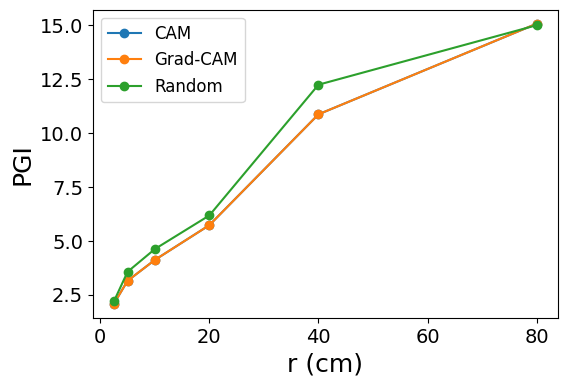}
        \caption{}\label{fig:class11_pgi}
    \end{subfigure}
    \vspace{5mm}
    \begin{subfigure}{0.3\textwidth}
        \includegraphics[width=\textwidth]{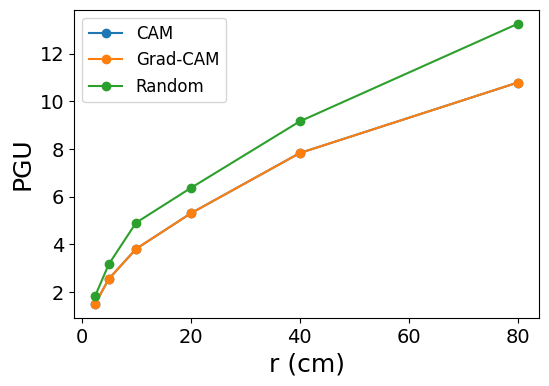}
        \caption{}\label{fig:class11_pgu}
    \end{subfigure}
    \begin{subfigure}{0.3\textwidth}
        \includegraphics[width=\textwidth]{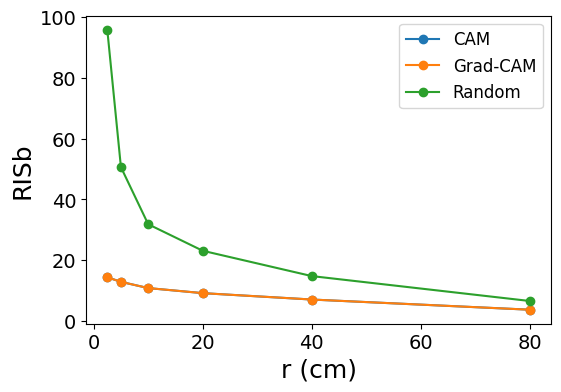}
        \caption{}\label{fig:class11_risb}
    \end{subfigure}
        \begin{subfigure}{0.3\textwidth}
        \includegraphics[width=\textwidth]{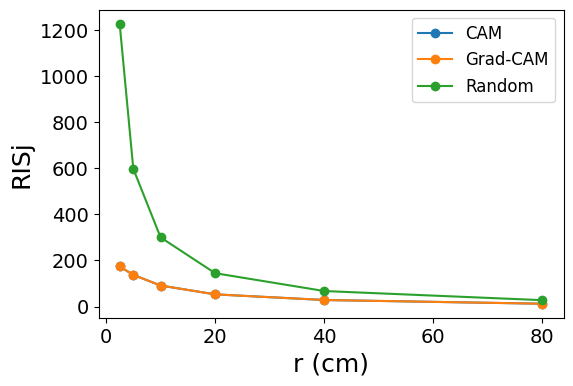}
        \caption{}\label{fig:class11_risj}
    \end{subfigure}
    \begin{subfigure}{0.3\textwidth}
        \includegraphics[width=\textwidth]{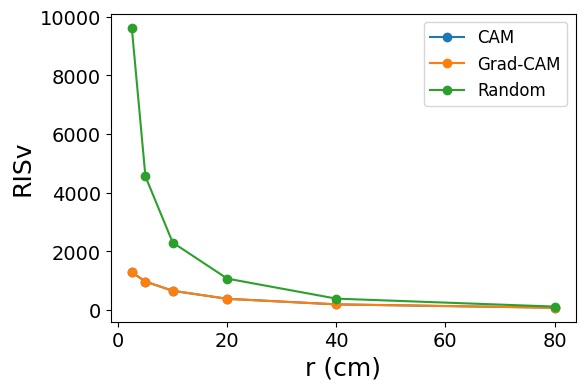}
        \caption{}\label{fig:class11_risv}
    \end{subfigure}
    \vspace{5mm}
    \begin{subfigure}{0.3\textwidth}
        \includegraphics[width=\textwidth]{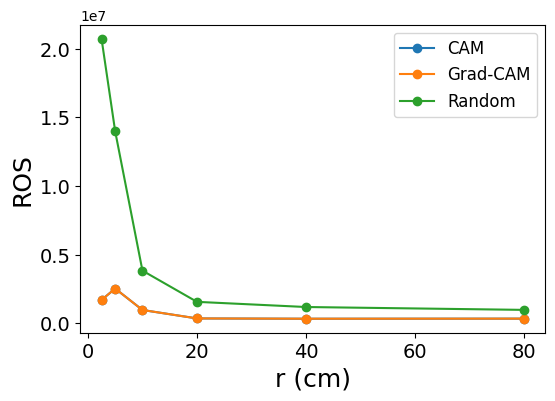}
        \caption{}\label{fig:class11_ros}
    \end{subfigure}
    \begin{subfigure}{0.3\textwidth}
        \includegraphics[width=\textwidth]{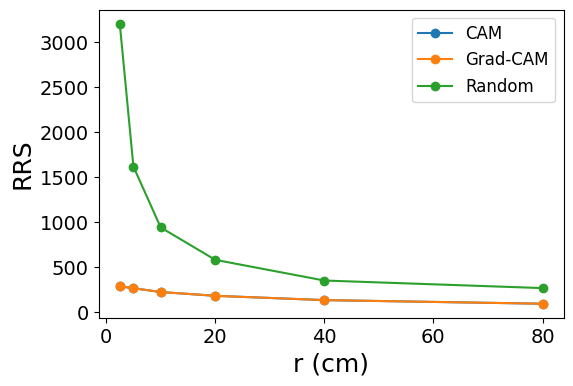}
        \caption{}\label{fig:class11_rrs}
    \end{subfigure}
    \caption{Evaluation metric outcomes for `Writing' (Class 11, i.e. the weakest class), showing CAM (blue), Grad-CAM (orange), and the random (green) methods, for 
    (\protect\subref{fig:class11_pgi}) PGI,
    (\protect\subref{fig:class11_pgu}) PGU,
    (\protect\subref{fig:class11_risb}) RISb,
    (\protect\subref{fig:class11_risj}) RISj,
    (\protect\subref{fig:class11_risv}) RISv,
    (\protect\subref{fig:class11_ros}) ROS, and
    (\protect\subref{fig:class11_rrs}) RRS.
    The $y$-axis measures the metric values, while the $x$-axis shows the perturbation magnitude. CAM and Grad-CAM graphs overlap due to extremely similar metric outcomes.}
    \label{fig:class11}
    \end{figure*}

\begin{figure*}
    \begin{subfigure}{0.3\textwidth}
        \includegraphics[width=\textwidth]{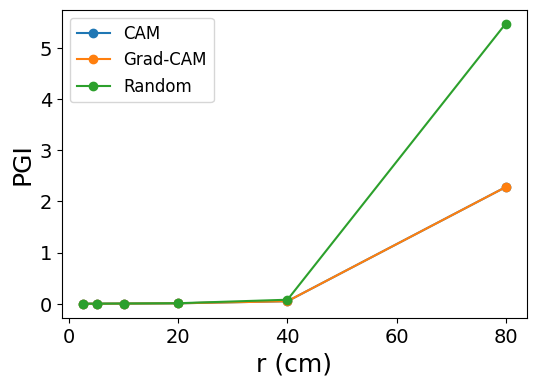}
        \caption{}\label{fig:class26_pgi}
    \end{subfigure}
    \vspace{5mm}
    \begin{subfigure}{0.3\textwidth}
        \includegraphics[width=\textwidth]{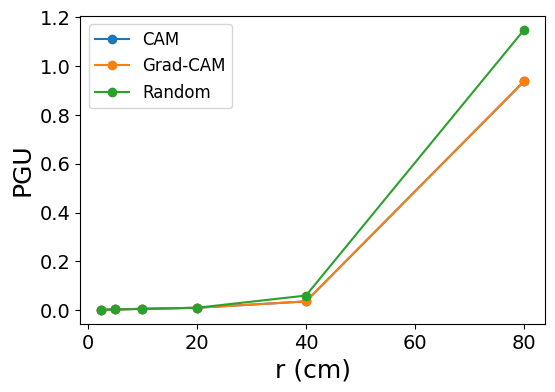}
        \caption{}\label{fig:class26_pgu}
    \end{subfigure}
    \begin{subfigure}{0.3\textwidth}
        \includegraphics[width=\textwidth]{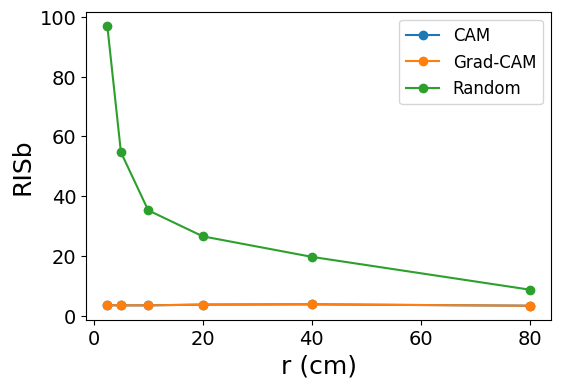}
        \caption{}\label{fig:class26_risb}
    \end{subfigure}
        \begin{subfigure}{0.3\textwidth}
        \includegraphics[width=\textwidth]{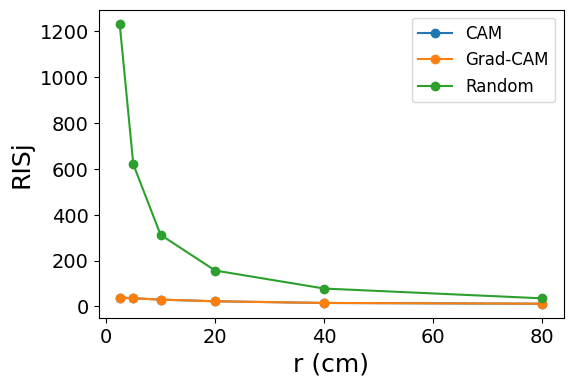}
        \caption{}\label{fig:class26_risj}
    \end{subfigure}
    \begin{subfigure}{0.3\textwidth}
        \includegraphics[width=\textwidth]{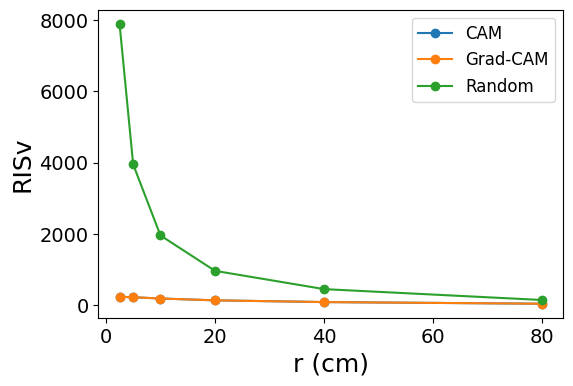}
        \caption{}\label{fig:class26_risv}
    \end{subfigure}
    \vspace{5mm}
    \begin{subfigure}{0.3\textwidth}
        \includegraphics[width=\textwidth]{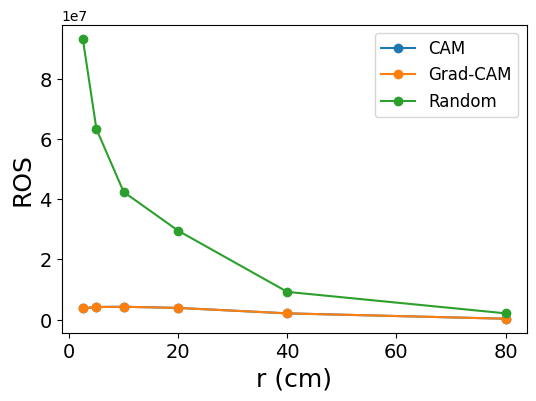}
        \caption{}\label{fig:class26_ros}
    \end{subfigure}
    \begin{subfigure}{0.3\textwidth}
        \includegraphics[width=\textwidth]{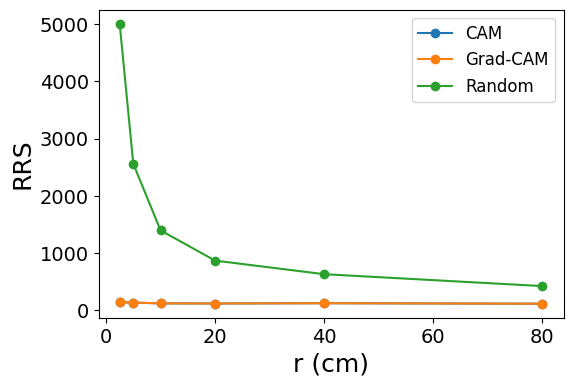}
        \caption{}\label{fig:class26_rrs}
    \end{subfigure}
    \caption{Evaluation metric outcomes for `Jump Up' (Class 26, i.e. the strongest class), showing CAM (blue), Grad-CAM (orange), and the random (green) methods, for 
    (\protect\subref{fig:class26_pgi}) PGI,
    (\protect\subref{fig:class26_pgu}) PGU,
    (\protect\subref{fig:class26_risb}) RISb,
    (\protect\subref{fig:class26_risj}) RISj,
    (\protect\subref{fig:class26_risv}) RISv,
    (\protect\subref{fig:class26_ros}) ROS, and
    (\protect\subref{fig:class26_rrs}) RRS.
    The $y$-axis measures the metric values, while the $x$-axis shows the perturbation magnitude. CAM and Grad-CAM graphs overlap due to extremely similar metric outcomes.}
    \label{fig:class26}
\end{figure*}

To help us gauge the reliability of the XAI evaluation metrics, we test them by slowly increasing the perturbation magnitude, as described in the Methods section. Figure~\ref{fig:class11} and~\ref{fig:class26} show the line graphs for each metric tests on class 11 and 26, respectively, comparing the different XAI methods. From hereon, we shall refer to class 26 as the strongest class, while class 11 will be referred to as the weakest class. For the exact numerical values of the results with confidence intervals, please refer to Appendix~\ref{appendix:class11data} and~\ref{appendix:class26data}. 

\subsection{Faithfulness}
The identical PGI and PGU values for CAM and Grad-CAM means both methods have the exact same ranking of features although the numerical attribution scores are not exactly the same. Unexpectedly, the random method appears to outperform both in PGI in the weakest class, except at $r=80cm$. Conversely, looking at the results for the strongest class in Figure~\ref{fig:class26} suggests equal PGI performance among all methods up to $r\leq5cm$, beyond which random seems to surpass the others. In essence, on a class where the model has the best classification performance, the PGI test aligns with expected outcomes only at higher values of $r$, where data distortion is significant, which is no longer consistent with the rule that $\mathbf{X}^{\prime} \in \mathcal{N}_{\mathbf{X}}$. Moreover, where model classification is weakest, PGI results consistently give unexpected outcomes. 

Analysis of the PGU results of the weakest class indicate that CAM and Grad-CAM outperform random as expected. In the strongest class, however, conformity to expected outcomes occur only when $r\geq40cm$, with random exhibiting higher values, while in lower perturbation magnitudes, the results seem to suggest that the three methods exhibit either comparable performance or that the random method has marginally higher PGU values. Therefore, PGU tests only meet expected outcomes primarily when there is weak model performance or when input perturbation is significant during strong model performance. These irregularities in both PGI and PGU suggest that the hypothesis of \textit{faithfulness}—minor perturbations causing meaningful prediction shifts—is not upheld for the EfficientGCN model. Using output predictions for gauging explanation fidelity proves unreliable in this context. 
\subsection{Stability}
Stability assessments for CAM and Grad-CAM yield nearly identical values, diverging only in less significant decimal places. This implies that despite both methods giving different raw scores, they tend to converge upon normalization. Stability test results, contrary to \textit{faithfulness}, demonstrate robustness against increased perturbation, consistently indicating the superiority of CAM and Grad-CAM over random in the two classes tested. Thus, stability testing affirms that input perturbations do not drastically alter explanations compared to the random baseline.

It can be observed in Figure~\ref{fig:class11_ros} and~\ref{fig:class26_ros} that ROS results register very high numerical values, with the $y$-axis scaled to $1 \times 10^7$. We inspected the individual terms in each of the ROS results and found that the cause for such high numbers is the extremely small denominator terms (typically less than 1). Since the denominator term of ROS is the difference in the original and perturbed predictions, it means that the change in the model's output probability is very small even when the perturbation magnitude is large. A small denominator, reflecting little changes in output probabilities even with substantial perturbations, corroborates the inefficacy of PGI and PGU tests in our context. These tests, which are reliant on shifts in prediction probabilities, fail to yield meaningful results in response to input perturbations, further supporting our hypothesis regarding the model's behavior under examination.

\section{Discussion and Conclusion}

Our research contributes to the understanding of explainable AI in the context of skeleton-based HAR, advancing the state-of-the-art by testing known metrics in this emerging domain and introducing a biomechanically-informed perturbation technique. A key finding from our experiments is that \textit{faithfulness}—a widely recognized XAI metric—may falter in certain models, such as the EfficientGCN. This finding serves as a caution to XAI practitioners when using an XAI metric that measures the reliability of XAI methods indirectly through the change in the model’s prediction probability. In contrast, \textit{stability}, which measures direct changes in explanations, emerged as a dependable metric. However, this leaves us with only a single metric which offers a limited view of an XAI method's efficacy, underscoring the need for developing or adapting additional testing approaches in this field. Our skeleton perturbation method, which leverages biomechanically precise modifications to evaluate explanation techniques, offers a promising framework for validating upcoming XAI metrics.

This study also identifies other gaps in XAI for ST-GCN-based HAR, which is an opportunity for future research directions. The nearly identical explanations produced by Grad-CAM and CAM when applied to EfficientGCN highlight a need for more diverse XAI techniques, such as adaptations of model agnostic methods like LIME \cite{ribeiro2016should} and SHAP \cite{lundberg2017unified} for this specific domain. Additionally, comparative studies of XAI metrics across various HAR models remain scarce, which could be valuable as a guide for model selection where explainability is as important as accuracy.

Lastly, our comparative analysis between CAM and Grad-CAM revealing negligible \textit{stability} differences suggests that neither method is superior; they are essentially equivalent. Yet, CAM’s use of static model weights obtained post-training means it demands less computational load compared to Grad-CAM, which needs gradient computation per data instance. This highlights CAM's suitability for large-scale data analysis. This consideration is especially pertinent for applications where computational efficiency is vital alongside accuracy and reliability.


\bibliographystyle{IEEEtran}
\bibliography{ref}

\begin{thebibliography}{10}
\providecommand{\url}[1]{#1}
\csname url@samestyle\endcsname
\providecommand{\newblock}{\relax}
\providecommand{\bibinfo}[2]{#2}
\providecommand{\BIBentrySTDinterwordspacing}{\spaceskip=0pt\relax}
\providecommand{\BIBentryALTinterwordstretchfactor}{4}
\providecommand{\BIBentryALTinterwordspacing}{\spaceskip=\fontdimen2\font plus
\BIBentryALTinterwordstretchfactor\fontdimen3\font minus
  \fontdimen4\font\relax}
\providecommand{\BIBforeignlanguage}[2]{{%
\expandafter\ifx\csname l@#1\endcsname\relax
\typeout{** WARNING: IEEEtran.bst: No hyphenation pattern has been}%
\typeout{** loaded for the language `#1'. Using the pattern for}%
\typeout{** the default language instead.}%
\else
\language=\csname l@#1\endcsname
\fi
#2}}
\providecommand{\BIBdecl}{\relax}
\BIBdecl

\bibitem{nguyen2016skeleton}
T.-N. Nguyen, H.-H. Huynh, and J.~Meunier, ``Skeleton-based abnormal gait
  detection,'' \emph{Sensors}, vol.~16, no.~11, p. 1792, 2016.

\bibitem{liu2021self}
C.~Liu, R.~Fu, Y.~Li, Y.~Gao, L.~Shi, and W.~Li, ``A self-attention augmented
  graph convolutional clustering networks for skeleton-based video anomaly
  behavior detection,'' \emph{Applied Sciences}, vol.~12, no.~1, p.~4, 2021.

\bibitem{guo2021attention}
J.~Guo, H.~Liu, X.~Li, D.~Xu, and Y.~Zhang, ``An attention enhanced
  spatial--temporal graph convolutional lstm network for action recognition in
  karate,'' \emph{Applied Sciences}, vol.~11, no.~18, p. 8641, 2021.

\bibitem{usman2022skeleton}
M.~Usman and J.~Zhong, ``Skeleton-based motion prediction: A survey,''
  \emph{Frontiers in Computational Neuroscience}, vol.~16, p. 1051222, 2022.

\bibitem{eu2021aiact}
\BIBentryALTinterwordspacing
E.~Commission, ``Proposal for a regulation of the european parliament and of
  the council laying down harmonised rules on artificial intelligence
  (artificial intelligence act) and amending certain union legislative acts,''
  European Commission, Proposal, 2021. [Online]. Available:
  \url{https://digital-strategy.ec.europa.eu/en/library/proposal-regulation-laying-down-harmonised-rules-artificial-intelligence-artificial-intelligence}
\BIBentrySTDinterwordspacing

\bibitem{yan2018spatial}
S.~Yan, Y.~Xiong, and D.~Lin, ``Spatial temporal graph convolutional networks
  for skeleton-based action recognition,'' in \emph{Proceedings of the AAAI
  conference on artificial intelligence}, vol.~32, 2018.

\bibitem{song2020stronger}
Y.-F. Song, Z.~Zhang, C.~Shan, and L.~Wang, ``Stronger, faster and more
  explainable: A graph convolutional baseline for skeleton-based action
  recognition,'' in \emph{proceedings of the 28th ACM international conference
  on multimedia}, 2020, pp. 1625--1633.

\bibitem{ghaleb2021skeleton}
E.~Ghaleb, A.~Mertens, S.~Asteriadis, and G.~Weiss, ``Skeleton-based
  explainable bodily expressed emotion recognition through graph convolutional
  networks,'' in \emph{2021 16th IEEE International Conference on Automatic
  Face and Gesture Recognition (FG 2021)}.\hskip 1em plus 0.5em minus
  0.4em\relax IEEE, 2021, pp. 1--8.

\bibitem{das2022gradient}
P.~Das and A.~Ortega, ``Gradient-weighted class activation mapping for spatio
  temporal graph convolutional network,'' in \emph{ICASSP 2022-2022 IEEE
  International Conference on Acoustics, Speech and Signal Processing
  (ICASSP)}.\hskip 1em plus 0.5em minus 0.4em\relax IEEE, 2022, pp. 4043--4047.

\bibitem{wang2021understanding}
H.~Wang, F.~He, Z.~Peng, T.~Shao, Y.-L. Yang, K.~Zhou, and D.~Hogg,
  ``Understanding the robustness of skeleton-based action recognition under
  adversarial attack,'' in \emph{Proceedings of the IEEE/CVF Conference on
  Computer Vision and Pattern Recognition}, 2021, pp. 14\,656--14\,665.

\bibitem{liu2020adversarial}
J.~Liu, N.~Akhtar, and A.~Mian, ``Adversarial attack on skeleton-based human
  action recognition,'' \emph{IEEE Transactions on Neural Networks and Learning
  Systems}, vol.~33, no.~4, pp. 1609--1622, 2020.

\bibitem{agarwal2022openxai}
C.~Agarwal, S.~Krishna, E.~Saxena, M.~Pawelczyk, N.~Johnson, I.~Puri,
  M.~Zitnik, and H.~Lakkaraju, ``Openxai: Towards a transparent evaluation of
  model explanations,'' \emph{Advances in Neural Information Processing
  Systems}, vol.~35, pp. 15\,784--15\,799, 2022.

\bibitem{zhou2021evaluating}
J.~Zhou, A.~H. Gandomi, F.~Chen, and A.~Holzinger, ``Evaluating the quality of
  machine learning explanations: A survey on methods and metrics,''
  \emph{Electronics}, vol.~10, no.~5, p. 593, 2021.

\bibitem{alvarez2018robustness}
D.~Alvarez-Melis and T.~S. Jaakkola, ``On the robustness of interpretability
  methods,'' \emph{arXiv preprint arXiv:1806.08049}, 2018.

\bibitem{shahroudy2016ntu}
A.~Shahroudy, J.~Liu, T.-T. Ng, and G.~Wang, ``Ntu rgb+ d: A large scale
  dataset for 3d human activity analysis,'' in \emph{Proceedings of the IEEE
  conference on computer vision and pattern recognition}, 2016, pp. 1010--1019.

\bibitem{song2022constructing}
Y.-F. Song, Z.~Zhang, C.~Shan, and L.~Wang, ``Constructing stronger and faster
  baselines for skeleton-based action recognition,'' \emph{IEEE transactions on
  pattern analysis and machine intelligence}, vol.~45, no.~2, pp. 1474--1488,
  2022.

\bibitem{tan2019efficientnet}
M.~Tan and Q.~Le, ``Efficientnet: Rethinking model scaling for convolutional
  neural networks,'' in \emph{International conference on machine
  learning}.\hskip 1em plus 0.5em minus 0.4em\relax PMLR, 2019, pp. 6105--6114.

\bibitem{otte2016accuracy}
K.~Otte, B.~Kayser, S.~Mansow-Model, J.~Verrel, F.~Paul, A.~U. Brandt, and
  T.~Schmitz-H{\"u}bsch, ``Accuracy and reliability of the kinect version 2 for
  clinical measurement of motor function,'' \emph{PloS one}, vol.~11, no.~11,
  p. e0166532, 2016.

\bibitem{zhou2016learning}
B.~Zhou, A.~Khosla, A.~Lapedriza, A.~Oliva, and A.~Torralba, ``Learning deep
  features for discriminative localization,'' in \emph{Proceedings of the IEEE
  conference on computer vision and pattern recognition}, 2016, pp. 2921--2929.

\bibitem{selvaraju2016grad}
R.~R. Selvaraju, A.~Das, R.~Vedantam, M.~Cogswell, D.~Parikh, and D.~Batra,
  ``Grad-cam: Why did you say that?'' \emph{arXiv preprint arXiv:1611.07450},
  2016.

\bibitem{ribeiro2016should}
M.~T. Ribeiro, S.~Singh, and C.~Guestrin, ``" why should i trust you?"
  explaining the predictions of any classifier,'' in \emph{Proceedings of the
  22nd ACM SIGKDD international conference on knowledge discovery and data
  mining}, 2016, pp. 1135--1144.

\bibitem{lundberg2017unified}
S.~M. Lundberg and S.-I. Lee, ``A unified approach to interpreting model
  predictions,'' \emph{Advances in neural information processing systems},
  vol.~30, 2017.

\end{thebibliography}

\onecolumn
\appendix
\section{Class 11 Metric Values}
\label{appendix:class11data}
\begin{table}[H]
\centering
\caption{Class 11 tabular data. $\uparrow$ indicates that higher values are better while $\downarrow$ indicates that lower values are optimal.}
\resizebox{\textwidth}{!}{%
\begin{tabular}{l|c|cccccccc}
\toprule
\textbf{Method} & \textbf{r (cm)} & \textbf{PGI} $ \uparrow $ & \textbf{PGU} $ \downarrow $ & \textbf{RISj} & \textbf{RISv} & \textbf{RISb} & \textbf{ROS} & \textbf{RRS} \\
\midrule
\multirow{6}{*}{CAM} & 2.5 & 2.080 ± 0.394 & 1.507 ± 0.268 & 174.133 ± 16.946 & 1288.370 ± 138.912 & 14.424 ± 1.202 & 1702553.834 ± 1116223.129 & 285.874 ± 46.673 \\
& 5 & 3.142 ± 0.514 & 2.547 ± 0.394 & 137.460 ± 11.303 & 968.090 ± 86.855 & 12.924 ± 0.964 & 2530301.100 ± 2226508.985 & 264.838 ± 33.044 \\
& 10 & 4.109 ± 0.569 & 3.807 ± 0.489 & 91.225 ± 6.774 & 654.886 ± 53.711 & 10.877 ± 0.741 & 971050.496 ± 676047.094 & 222.611 ± 20.795 \\
& 20 & 5.722 ± 0.616 & 5.294 ± 0.541 & 52.854 ± 3.473 & 380.917 ± 30.189 & 9.210 ± 0.556 & 355153.157 ± 272318.174 & 180.486 ± 11.691 \\
& 40 & 10.859 ± 0.811 & 7.826 ± 0.549 & 28.373 ± 2.139 & 190.976 ± 18.548 & 7.112 ± 0.416 & 334731.175 ± 221923.029 & 132.782 ± 8.488 \\
& 80 & 15.059 ± 1.023 & 10.787 ± 0.679 & 12.199 ± 1.408 & 76.734 ± 10.911 & 3.764 ± 0.330 & 340629.702 ± 291218.222 & 92.569 ± 8.165 \\
\addlinespace
\multirow{6}{*}{Grad-CAM} & 2.5 & 2.080 ± 0.394 & 1.507 ± 0.268 & 174.133 ± 16.946 & 1288.371 ± 138.912 & 14.424 ± 1.202 & 1702554.201 ± 1116223.411 & 285.874 ± 46.673 \\
& 5 & 3.142 ± 0.514 & 2.547 ± 0.394 & 137.460 ± 11.303 & 968.090 ± 86.855 & 12.924 ± 0.964 & 2530301.513 ± 2226509.277 & 264.838 ± 33.044 \\
& 10 & 4.109 ± 0.569 & 3.807 ± 0.489 & 91.225 ± 6.774 & 654.886 ± 53.711 & 10.877 ± 0.741 & 971050.337 ± 676047.046 & 222.611 ± 20.795 \\
& 20 & 5.722 ± 0.616 & 5.294 ± 0.541 & 52.854 ± 3.473 & 380.917 ± 30.189 & 9.210 ± 0.556 & 355153.184 ± 272318.220 & 180.486 ± 11.691 \\
& 40 & 10.859 ± 0.811 & 7.826 ± 0.549 & 28.373 ± 2.139 & 190.976 ± 18.548 & 7.112 ± 0.416 & 334731.302 ± 221923.109 & 132.782 ± 8.488 \\
& 80 & 15.059 ± 1.023 & 10.787 ± 0.679 & 12.199 ± 1.408 & 76.734 ± 10.911 & 3.764 ± 0.330 & 340629.723 ± 291218.252 & 92.569 ± 8.165 \\
\addlinespace
\multirow{6}{*}{Random} & 2.5 & 2.190 ± 0.410 & 1.831 ± 0.306 & 1228.105 ± 43.441 & 9627.812 ± 502.591 & 95.762 ± 3.630 & 20728354.210 ± 14139843.294 & 3205.744 ± 333.826 \\
& 5 & 3.559 ± 0.577 & 3.165 ± 0.445 & 598.011 ± 21.123 & 4555.435 ± 238.309 & 50.612 ± 1.843 & 13986453.073 ± 9801817.526 & 1617.393 ± 160.022 \\
& 10 & 4.613 ± 0.664 & 4.905 ± 0.542 & 300.288 ± 11.303 & 2300.006 ± 112.759 & 31.848 ± 1.120 & 3835957.191 ± 2481844.039 & 940.518 ± 87.484 \\
& 20 & 6.173 ± 0.643 & 6.357 ± 0.596 & 144.791 ± 5.982 & 1069.703 ± 56.382 & 23.135 ± 0.976 & 1563644.666 ± 1072148.002 & 580.579 ± 46.894 \\
& 40 & 12.232 ± 0.820 & 9.155 ± 0.654 & 67.294 ± 3.949 & 386.528 ± 35.895 & 14.813 ± 0.768 & 1187301.059 ± 987974.000 & 350.415 ± 26.835 \\
& 80 & 14.989 ± 1.008 & 13.253 ± 0.852 & 27.610 ± 2.828 & 115.425 ± 19.128 & 6.621 ± 0.418 & 978432.004 ± 1211946.037 & 266.365 ± 27.954 \\
\bottomrule
\end{tabular}%
}
\end{table}

\section{Class 26 Metric Values}
\label{appendix:class26data}
\begin{table}[H]
\centering
\caption{Class 26 tabular data. $\uparrow$ indicates that higher values are better while $\downarrow$ indicates that lower values are optimal.}
\resizebox{\textwidth}{!}{%
\begin{tabular}{l|c|cccccccc}
\toprule
\textbf{Method} & \textbf{r (cm)} & \textbf{PGI} $ \uparrow $ & \textbf{PGU} $ \downarrow $ & \textbf{RISj} & \textbf{RISv} & \textbf{RISb} & \textbf{ROS} & \textbf{RRS} \\
\midrule
\multirow{6}{*}{CAM} & 2.5 & 0.002 ± 0.001 & 0.002 ± 0.001 & 38.783 ± 1.685 & 241.329 ± 10.033 & 3.504 ± 0.112 & 3793568.715 ± 599734.044 & 137.439 ± 4.602 \\
& 5 & 0.003 ± 0.001 & 0.003 ± 0.001 & 35.563 ± 1.469 & 226.272 ± 8.821 & 3.447 ± 0.103 & 4244155.156 ± 614303.198 & 130.751 ± 3.800 \\
& 10 & 0.005 ± 0.002 & 0.006 ± 0.003 & 29.652 ± 1.165 & 191.308 ± 7.224 & 3.452 ± 0.110 & 4307133.768 ± 667017.130 & 116.767 ± 3.276 \\
& 20 & 0.010 ± 0.005 & 0.010 ± 0.004 & 22.343 ± 0.782 & 140.125 ± 5.027 & 3.698 ± 0.108 & 3926629.242 ± 573923.634 & 114.482 ± 3.087 \\
& 40 & 0.051 ± 0.040 & 0.036 ± 0.014 & 15.016 ± 0.481 & 91.628 ± 3.543 & 3.793 ± 0.111 & 2119077.311 ± 355383.073 & 119.608 ± 2.883 \\
& 80 & 2.281 ± 0.299 & 0.938 ± 0.140 & 11.679 ± 0.347 & 45.246 ± 2.827 & 3.324 ± 0.090 & 339790.687 ± 75217.805 & 110.699 ± 2.141 \\
\addlinespace
\multirow{6}{*}{Grad-CAM} & 2.5 & 0.002 ± 0.001 & 0.002 ± 0.001 & 38.783 ± 1.685 & 241.329 ± 10.033 & 3.504 ± 0.112 & 3793569.048 ± 599734.059 & 137.439 ± 4.602 \\
& 5 & 0.003 ± 0.001 & 0.003 ± 0.001 & 35.563 ± 1.469 & 226.272 ± 8.821 & 3.447 ± 0.103 & 4244156.522 ± 614303.356 & 130.751 ± 3.800 \\
& 10 & 0.005 ± 0.002 & 0.006 ± 0.003 & 29.652 ± 1.165 & 191.308 ± 7.224 & 3.452 ± 0.110 & 4307133.975 ± 667017.145 & 116.767 ± 3.276 \\
& 20 & 0.010 ± 0.005 & 0.010 ± 0.004 & 22.343 ± 0.782 & 140.125 ± 5.027 & 3.698 ± 0.108 & 3926629.059 ± 573923.586 & 114.482 ± 3.087 \\
& 40 & 0.051 ± 0.040 & 0.036 ± 0.014 & 15.016 ± 0.481 & 91.628 ± 3.543 & 3.793 ± 0.111 & 2119077.265 ± 355383.079 & 119.608 ± 2.883 \\
& 80 & 2.281 ± 0.299 & 0.938 ± 0.140 & 11.679 ± 0.347 & 45.246 ± 2.827 & 3.324 ± 0.090 & 339790.661 ± 75217.791 & 110.699 ± 2.141 \\
\addlinespace
\multirow{6}{*}{Random} & 2.5 & 0.002 ± 0.001 & 0.002 ± 0.001 & 1233.418 ± 22.376 & 7891.926 ± 206.386 & 97.067 ± 1.616 & 93264563.501 ± 12716437.799 & 5008.974 ± 133.687 \\
& 5 & 0.003 ± 0.002 & 0.003 ± 0.001 & 619.974 ± 10.867 & 3948.918 ± 101.998 & 54.711 ± 0.727 & 63286627.152 ± 9322027.632 & 2557.063 ± 64.789 \\
& 10 & 0.006 ± 0.002 & 0.005 ± 0.002 & 312.310 ± 5.332 & 1958.503 ± 50.638 & 35.273 ± 0.333 & 42392862.818 ± 6446239.434 & 1393.590 ± 32.140 \\
& 20 & 0.012 ± 0.005 & 0.010 ± 0.004 & 156.615 ± 2.673 & 966.698 ± 25.445 & 26.557 ± 0.193 & 29527933.778 ± 4113487.016 & 864.778 ± 18.918 \\
& 40 & 0.080 ± 0.055 & 0.060 ± 0.027 & 77.913 ± 1.346 & 454.485 ± 13.903 & 19.684 ± 0.241 & 9244726.238 ± 1352133.114 & 627.195 ± 13.200 \\
& 80 & 5.474 ± 0.334 & 1.149 ± 0.176 & 35.110 ± 0.858 & 149.132 ± 7.404 & 8.684 ± 0.145 & 2129144.307 ± 407137.423 & 419.232 ± 8.483 \\
\bottomrule
\end{tabular}%
}
\end{table}

\end{document}